\pdfoutput=1

\documentclass[11pt]{article}
\usepackage[hyperref]{naacl2021}

\usepackage{times}
\usepackage{booktabs}       
\usepackage{latexsym}
\usepackage{hyperref}       
\usepackage{booktabs}       
\usepackage{amsfonts}       
\usepackage[inline,shortlabels]{enumitem}
\usepackage{todonotes}
\usepackage{amsmath}
\usepackage{siunitx}
\usepackage{amssymb}
\usepackage{threeparttable}
\usepackage{natbib}
\usepackage{flushend}
\usepackage{color}

\makeatletter
\let\UrlSpecialsOld\UrlSpecials
\def\UrlSpecials{\UrlSpecialsOld\do\/{\Url@slash}\do\_{\Url@underscore}}%
\def\Url@slash{\@ifnextchar/{\kern-.11em\mathchar47\kern-.2em}%
    {\kern-.0em\mathchar47\kern-.08em\penalty\UrlBigBreakPenalty}}
\makeatother

\usepackage{microtype}

\newcommand{\pretraining}{$\rightarrow$ $\pi$}
\newcommand{\finetuning}{$\rightarrow$ $\varphi$}
\newcommand{\freeze}{$\rightarrow$ $\varnothing$}

\newcommand{\eg}{e.g., }
\newcommand{\ie}{i.e., }
\newcommand{\figref}[1]{Fig.~\ref{#1}}    
\newcommand{\tabref}[1]{Table~\ref{#1}}
\newcommand{\Tabref}[1]{Table~\ref{#1}}
\newcommand{\secref}[1]{Section~\ref{#1}}

\title{A Million Tweets Are Worth a Few Points:\\
Tuning Transformers for Customer Service Tasks}

\author{Amir Hadifar$^{\dag}$ \qquad Sofie Labat$^{\ddag}$ \qquad Véronique Hoste$^{\ddag}$ \\ \textbf{Chris Develder}$^{\dag}$ \qquad \textbf{Thomas Demeester}$^{\dag}$ \\

	$^{\dag}$ IDLab, Dept.~of Information Technology, Ghent University - imec, Belgium \\
	$^{\ddag}$ LT3, Dept.~of Translation, Interpreting and Communication, Ghent University, Belgium \\
}
	
\begin{document}
\maketitle
\begin{abstract}
In online domain-specific customer service applications, many companies struggle to deploy advanced NLP models successfully, due to the limited availability of and noise in their datasets. While prior research demonstrated the potential of migrating large open-domain pretrained models for domain-specific tasks, the appropriate (pre)training strategies have not yet been rigorously evaluated in such social media customer service settings, especially under multilingual conditions. We address this gap by 
\begin{enumerate*}[(i)]
    \item collecting a multilingual social media corpus containing customer service conversations (865k tweets), 
    \item comparing various pipelines of pretraining and finetuning approaches,
    \item applying them on 5 different end tasks.
\end{enumerate*}
We show that pretraining a generic multilingual transformer model on our in-domain dataset, before finetuning on specific end tasks, consistently boosts performance, especially in non-English settings.\footnote{Dataset and code available at \url{https://github.com/hadifar/customerservicetasks}}

\end{abstract}

\section{Introduction}
\label{sec:intro}

Online platforms and social media are increasingly important as communication channels in various companies' customer relationship management (CRM). 
To ensure effective, qualitative and timely customer service, Natural Language Processing (NLP) can assist by providing insights 
to optimize customer interactions, but also in real-time tasks:
\begin{enumerate*}[(i)]
    \item detect emotions~\cite{gupta-etal-2010-emotion},
    \item categorize or prioritize customer tickets~\cite{molino2018cota},
    \item aid in virtual assistants through natural language understanding and/or generation~\cite{cui-etal-2017-superagent}, etc.
\end{enumerate*}

Despite this NLP progress for CRM, often small and medium-sized companies (SMEs) struggle with applying such recent technology due to the limited size, noise and imbalance in their datasets. General solutions to such challenges are transfer learning strategies
\cite{Ruder2019Neural}: \emph{feature extraction} uses frozen model parameters after pretraining on an external (larger) training corpus, while \emph{finetuning}
continues training on the smaller in-domain corpus. In the large body of work adopting such strategies (\eg \citealt{pan2009survey}), little effort has been put into addressing specific CRM use cases that need to rely on social media data that is noisy, possibly multilingual, and domain-specific for a given company.

In this paper, we analyze the possibilities and limitations of transfer learning for a number of CRM tasks, following up on the findings of \citet{gururangan2020don} who demonstrate gains from progressive finetuning on in-domain and task-specific monolingual data. Specifically, our contributions are that we
\begin{enumerate*}[(1)]
    \item collect a multilingual corpus of 275k Twitter conversations, comprising 865k tweets between customers and companies in 4 languages (EN, FR, DE, NL), covering distinct sectors (telecom, public transport, airline) (\secref{sec:data_construction});
    \item rigorously compare combinations of pretraining and finetuning strategies (\secref{sec:methodology}) on 5 different CRM tasks (\secref{sec:tasks_datasets}), including prediction of complaints, churn, subjectivity, relevance, and polarity; and
    \item provide empirical results (\secref{sec:results}). 
\end{enumerate*}
We find that additional pretraining on a moderately sized in-domain corpus, before task-specific finetuning, contributes to overcoming the lack of a large multilingual domain-specific language model. Its effect is much stronger than consecutive finetuning on smaller datasets for related end tasks. Furthermore, our experimental results show that when pretrained models are used in feature extraction mode, they struggle to beat well-tuned classical baselines.

\section{Related Work}

A wide range of NLP research has been devoted to customer services.
\citet{hui2000data} employed data mining techniques to extract features from a customer service database for decision support and machine fault diagnosis. \citet{gupta2011extracting} extracted a set of sentiment and syntactic features from tweets for customer problem identification tasks.
\citet{molino2018cota} introduced the Customer Obsession Ticket Assistant for ticket resolution, using feature engineering techniques and encoder-decoder models.

Highly popular pretrained language models, such as BERT
\cite{devlin2019bert}, have also been explored for different customer service tasks:
\citet{hardalov2019machine} considered re-ranking candidate answers in
chatbots, while \citet{deng2020semi} proposed BERT-based topic prediction for incoming customer requests.
Although the performance gains obtained by pretraining language models are well-established, they need further exploration in terms of multilinguality.
Some studies \cite{pires2019multilingual, karthikeyan2019cross, wu2019emerging} have investigated the transferability of multilingual models on different tasks, but they do not consider the effect of progressive pretraining on a smaller and less diverse multilingual corpus, as we will do.

\section{Methodology}
\label{sec:methodology}

\subsection{Architecture}
\label{sec:methodology_arc}
We selected some of the most popular publicly available pretrained language models to explore transfer learning properties for CRM classification tasks:
RoBERTa \cite{liu2019roberta}, XLM \cite{conneau2019unsupervised}, and BERTweet \cite{bertweet}. 
These models are pretrained on the English Wikipedia and BookCorpus \cite{zhu2015aligning}, CommonCrawl in 100 languages, and 850M English tweets, respectively. 
The XLM and BERTweet pretraining procedure is based on RoBERTa, which itself is a transformer-based Masked Language Model \citep[MLM;][]{devlin2019bert}. All of these models
require a different classifier `head' for each target task to estimate the probability of a class label.

\subsection{Transfer Strategies}
\label{sec:methodology_train}

We adopt a straightforward approach to transfer learned representations:
we continue pretraining the considered transformer models on a 4-lingual corpus of customer service Twitter conversations (see \secref{sec:data_construction}), \ie the overall domain of all considered sub-tasks. After that, we apply additional adaptation for cross-lingual transfer (\secref{mono_multi}), as well as cross-task transfer (\secref{task_trasferability}).

The following notations are used throughout the rest of this paper to describe pretraining stages:
\begin{itemize}[nosep]
  \setlength\itemsep{0em}
  \item $\pi$  --  further \textbf{p}retraining the original MLM on our 4-lingual tweet conversation corpus.
  \item $\varphi$ -- 
   \textbf{fi}netuning the pretrained model extended with the MLP classifier on the target task 
  
  \item $\varnothing$ -- freezing the pretrained model (\ie feature extraction mode), only training the top classifier on the target task.
\end{itemize}

We thus indicate several multistage procedures: \eg XLM$\rightarrow{\pi}\rightarrow{\varphi}$ indicates that the XLM model is further pretrained on the in-domain tweet corpus, followed by finetuning on the end task.

\section{Experimental Setup}
We focus our experiments on text classification problems that are commonly dealt with by customer service teams. 
First, we describe our Twitter conversation corpus used for in-domain finetuning (\secref{sec:data_construction}), then we 
introduce the target tasks and corresponding datasets
(\secref{sec:tasks_datasets}).
For most target tasks, we hold out 10\% of the data for testing, while the remaining part is used for training.
We then utilize 10-fold cross-validation on the training data to select optimal hyper-parameters for each end task.
When the dataset comes with a predefined train-test split, we keep that. 
For the pretrained transformer models (RoBERTa, XLM, BERTweet), we use the publicly available `base' versions.

\subsection{Twitter Conversation Corpus}
\label{sec:data_construction}
Our corpus for in-domain pretraining was crawled
using Twitter's API.\footnote{\url{https://developer.twitter.com/en/docs}} The collected dataset is small compared to the original language models' data, but still larger than most corpora which SMEs have at their disposal. As such, it represents an easily collectable customer service dataset that SMEs can leverage to boost models on their own data. 

The tweets were gathered between May and October 2020. 
We started by gathering a list of 104 companies, all active on Twitter, in the sectors of telecommunication, public transportation, and airlines. We aimed for four different languages (English, French, Dutch, German).

We preprocessed the data by removing conversations not covering at least one client/company interaction, or containing undefined languages. We further converted usernames and links into the special tokens @USER and @HTTP\_URL, respectively, and converted emojis into corresponding strings.\footnote{We used \url{https://github.com/carpedm20/emoji} to convert emojis} The resulting corpus contains 865k tweets over 275k conversations in the four target languages (see \tabref{tb:Dataset}). 

\begin{figure}
  \centering
  \includegraphics[trim={1cm 1cm 1cm 0cm},scale=0.50]{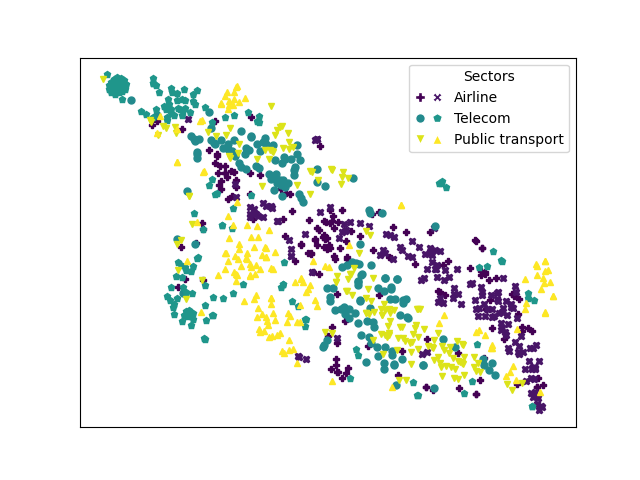}
  \caption{2D visualization of the first hidden-state representations of the XLM on randomly sampled tweets for six companies, in three different sectors using t-SNE \cite{van2008visualizing}. Each color stands for a different sector, while the marker types represent the different companies, and each data point corresponds to a single operator tweet (covering multiple languages).
  }
  \label{fig_tsne}
\end{figure}

Even though our corpus contains data from different sectors, we noticed that the dialogue flow, customer intents, and structure of conversations are fairly comparable across the target sectors (cf.~\figref{fig_tsne}). Examples of often recurring types of tweets are expressions of gratitude towards customers, requests for information, or typical ways to reply to complaints. 
Hence, we expect this corpus to be useful not only for companies that fall under one of the included sectors, but also for
other companies that provide customer sewrvices over tweets.

\begin{table}[t!]
\footnotesize
\begin{center}
\begin{tabular}{ccc}
\toprule
Language & $|\text{convs}|$ & $|\text{tweets}|$ \\
\midrule
English   & 135.1k   &	406.3k \\
French     & 60.9k    &	212.9k \\
Dutch     &	45.6k   &	141.0k  \\
German     &	33.0k    &	104.5k \\
\midrule
All     &	274.6k    &	864.7k   \\
\bottomrule
\end{tabular}
\end{center}
\caption{\label{tb:Dataset} Number of collected conversations ($|\text{convs}|$) and tweets ($|\text{tweets}|$) for each language. 
}
\end{table}

\begin{table*}[htbp]
    \centering
    \renewcommand{\arraystretch}{1.2}
    \footnotesize
    \label{main_tab}
    \begin{tabular}{l cc cc cc cc cc cc}
    \toprule
     & \multicolumn{2}{c}{\textbf{Complaint-2}} 
     & \multicolumn{2}{c}{\textbf{Complaint-R}}
     & \multicolumn{2}{c}{\textbf{Churn}}
     & \multicolumn{2}{c}{\textbf{Subjectivity}}
     & \multicolumn{2}{c}{\textbf{Relevance} }
     & \multicolumn{2}{c}{\textbf{Polarity} }
    \\
    & \multicolumn{2}{c}{(English)} & \multicolumn{2}{c}{(French)} &
    \multicolumn{2}{c}{(English)} & \multicolumn{2}{c}{(Dutch)} &
    \multicolumn{2}{c}{(German)} & \multicolumn{2}{c}{(German)}  
    \\
    \cmidrule(r{3pt}){2-3}
    \cmidrule(l{3pt}r{3pt}){4-5}
    \cmidrule(l{3pt}r{3pt}){6-7}
    \cmidrule(l{3pt}r{3pt}){8-9}
    \cmidrule(l{3pt}r{3pt}){10-11}
    \cmidrule(l{3pt}){12-13}

    \textbf{Model} &
    \textbf{ACC}&  \textbf{F1} &
    \textbf{ACC} & \textbf{F1} &
    \textbf{ACC} & \textbf{F1} &
    \textbf{ACC} & \textbf{F1} &
    
    \textbf{F1$_{syn.}$} & 
    \textbf{F1$_{dia.}$} &
    \textbf{F1$_{syn.}$} & 
    \textbf{F1$_{dia.}$} \\

    \midrule
     Majority-class   & 72.4 & 42.0  & 56.0 & 35.8 & 78.4 & 43.9 & 55.0 & 35.5 & 81.6 &  83.9 & 65.6 & 67.2   \\

     LR (tf-idf)    & 83.5 & 77.0 & 57.5 & 57.4 & 85.1 & 71.7 & 71.6 & 70.9  & 88.4 &  87.7 & 71.1 &  70.4   \\

    SVM (tf-idf)  & 84.4 & 80.2 & 59.0 & 58.8 & 87.3 & 80.1 & 71.7 & 71.0  & 90.4 &  88.8 & 74.8 &  72.8 \\ 
    
    Reference & 82.0$^{[1]}$ & 62.7$^{[1]}$  & - & - & - & 78.3$^{[2]}$  & - & - & 85.2$^{[3]}$ &  86.8$^{[3]}$ & 66.7$^{[3]}$ & 69.4$^{[3]}$  \\
     
    \midrule
    
    BERTweet \freeze   & 80.5 & 71.6  & - & - & 79.3 & 55.2  & - & - & - & - & - & -  \\
    
    BERTweet \finetuning    &  \textbf{90.0} & \textbf{86.1}  & - & - & \textbf{93.0} & \textbf{90.0}  & - & - & - & - & - & - \\
    
    RoBERTa \freeze    & 77.9 & 74.5  & - & - & 78.3 & 59.7  & - & - & - & - & - & - \\
    
    RoBERTa \finetuning    & 87.5 & 85.1  & - & - & 88.4 & 84.8  & - & - & - & - & - & - \\

    XLM \freeze    & 76.2 & 61.6 & 44.0 & 30.5 & 61.6  & 55.7 & 63.8 & 62.4 & 83.1 & 84.7 & 64.5 & 66.8  \\

    XLM \finetuning   & 85.4 & 83.4  & 54.0 & 46.2 & 84.1 & 75.3 & 73.4 & 72.9 & 91.6 & 91.7 & 76.1 & 73.5  \\

    XLM \pretraining \freeze    & 81.8 & 76.8 & 56.5 & 54.1 & 79.7 & 66.0 & 71.6 & 71.1 & 84.4 & 85.3 & 65.1 & 68.0  \\

    XLM \pretraining \finetuning    & 86.9 & 82.7 & \textbf{62.0} & \textbf{61.9} & 87.8 & 83.7  & \textbf{74.6} & \textbf{74.2} & \textbf{92.7} & \textbf{92.5} & \textbf{78.7} & \textbf{76.1} \\
    
    \bottomrule

    \end{tabular}
        \caption{\label{tb:Results} Classification results (accuracy ACC and F1-score) on CRM tasks using pretrained language models with two settings for pretraining: Feature extraction (\freeze) and finetuning (\finetuning). Missing values (`-') are due to unavailable reference scores, or a language mismatch between model and task. \\ 
        $^{[1]}$ \citet{RigautAntoine2015final}, $^{[2]}$ \citet{amiri2015target}, $^{[3]}$ \citet{germevaltask2017}
    }
    \label{tb:classification_res}
\end{table*}

\subsection{CRM Tasks and Datasets}
\label{sec:tasks_datasets}

\noindent\textbf{Complaint Prediction --} Timely complaint detection is of utmost importance to organizations, as it can improve their relationship with customers and prevent customer churns. \citet{complaints2019acl} and \citet{RigautAntoine2015final} proposed two datasets for identifying complaints on social media which contain 3,499 and 5,143 instances, respectively. The former (\textbf{Complaint-2}) covers two types of companies (airline companies and telecommunication), while the latter (\textbf{Complaint-9}) consists of data from nine domains such as food, car, software, etc. 

Both datasets are in English.
To experiment with cross-lingual tuning for complaint prediction, we use the French complaint dataset for railway companies from (\textbf{Complaint-R}; \citealt{railway8677442}). Since all their 201 conversations are labeled as complaints, for training, we complemented them with negative sampling from French railway conversations in our own Twitter corpus. For testing, we annotated 200 held-out conversations.

\noindent\textbf{Churn prediction --} Customer churn implies that a customer stops using a company's service, negatively impacting its growth. Churn prediction is cast as a binary classification task (churn or non-churn) on any input text.  We utilize the data provided by~\citet{amiri2015target} with tweets from three telecommunication brands, resulting in a corpus of 4,339 labelled English tweets.

\noindent\textbf{Subjectivity Prediction --} Detecting subjectivity in conversations is a key task for companies to efficiently address negative customer feelings or reward loyal customers. It may also serve as a filtering task for more fine-grained tasks such as emotion identification.
We annotated 8,174 Dutch conversations from our Twitter corpus (\secref{sec:data_construction}). 
A dialogue is judged ``subjective'' if at least one of the customer turns contains emotions (explicit or implicit), and otherwise ``objective''.

\noindent\textbf{Relevance Prediction --} The goal of this task is to determine whether an incoming text is relevant for further processing or not.

We use data from GermEval 2017 (Task A) which contains over 28k short length messages from various social media and web sources
on the German public train operator Deutsche Bahn~\cite{germevaltask2017}. For this dataset, the evaluation is measured on two evaluation sets: one collected from the same time period as the training and development set (viz.~synchronic), and another one containing data from a later time period (viz.~diachronic).

\noindent\textbf{Polarity Prediction --} For this task, a system has to classify the sentiment that resides in a given text fragment according to polarity (positive, negative, or neutral). Polarity prediction has often been applied on reviews, by predicting the attitude or sentiment of a reviewer with respect to some topic. We use the GermEval 2017 (Task B) dataset~\cite{germevaltask2017} (cf. supra) to analyze the polarity of the Deutsche Bahn customers' feedback. We also use the polarity dataset from \citet{sander2011} (\textbf{Sanders}).

\section{Results and Discussion}
\label{sec:results}

We now present our findings for two finetuning scenarios: transfer across languages and across tasks.
\secref{mono_multi} investigates the effect of unsupervised multilingual pretraining. \secref{task_trasferability} then explores 
how to further improve by finetuning the pretrained language models on similar tasks.

\subsection{Language Transferability}
\label{mono_multi}
We compare the pretrained transformer experiments with the following baselines: majority-class (to get an idea of class imbalance), logistic regression (LR) and support vector machine (SVM) with tf-idf features. 
For the three transformer models, we compare the feature extraction setting (\freeze) with finetuning (\finetuning) on the target task. On the multilingual XLM, we measure the impact of first pretraining (\pretraining) on our multilingual tweet corpus, after which both transfer settings are again tested on the target tasks. \Tabref{tb:classification_res} reports the results (in terms of accuracy and F1 scores), including scores from literature when available (`Reference'). It should be noted that the reference scores are not state-of-the-art, but they are the scores communicated in the original dataset papers.

Only for the English tasks (Complaint-2 and Churn), results for BERTweet and RoBERTa are reported. The monolingual tweet-based model BERTweet outperforms all other models when finetuned on these tasks. 
Although a large domain-specific mono-lingual language model seems a fine choice, it may not be available for other languages. We therefore investigate the impact of a multilingual generic model (XLM was not specifically pretrained on tweets), and the impact of additional finetuning on our dedicated twitter corpus.

In general, transformer models finetuned on the end task strongly outperform frozen ones. For the non-English tasks, the model XLM\freeze~with the frozen XLM encoder shows weak performance, in some cases below the baselines. The model XLM\finetuning~finetuned on the end task performs better.
For the non-English tasks, the XLM model pretrained on our Twitter corpus and finetuned on the tasks (XLM\pretraining\finetuning) in all cases
outperforms the finetuned XLM by a few percentage points and the baselines by an even larger margin. The performance differences between XLM\finetuning\space and XLM\pretraining\finetuning\space clearly underscore the importance of in-domain multilingual pretraining. Furthermore, the results of XLM\pretraining\finetuning\space for the English tasks suggest that additional pretraining on a moderately small, in-domain dataset can make the performance of the multilingual XLM model comparable to the monolingual RoBERTa.

Another promising observation is that the hyper-tuned classical baselines, such as SVM, are strong competitors compared to \emph{frozen} language models, especially on tasks that are highly sensitive to domain-specific features. For instance, for churn prediction, keywords such as ‘switch to’, ‘quit’ and ‘change provider’ can easily be triggered by the SVM, while \emph{frozen} pretrained models have not learned to identify these features. This finding might be helpful to achieve better insight into the operational aspects of \emph{frozen} neural models compared to simple classical approaches.

As a side result (not explicitly included in this work)
we found that the multistage pretraining (XLM\pretraining) leads to better performance when incorporating multiple languages compared to a single language. The performance drops especially when training data from a single language (\eg Dutch) is fed into the model, which is then evaluated on other languages (\eg English). 

\subsection{Task Transferability}
\label{task_trasferability}
We now investigate to what extent representations tuned on a related task can help for a given target task. In particular, Complaint-9 is the end task, and we compare the effect of finetuning on the end task only, vis-à-vis first finetuning on a related task and then on the end task. For the related task, we experiment with Complaint-2 and Sanders, as shown in \tabref{tb:task_transferability}. We observe that there seems to be no clear merit in the additional finetuning step on a small related end task. Pretraining on our larger Twitter corpus, however, still increases effectiveness.

\begin{table}[ht!]
\begin{center}
\footnotesize

\begin{tabular}{lcccccc}
\toprule
\textbf{Test Dataset} & 
\multicolumn{3}{c}{\textbf{Complaint-9}}
\\\midrule
\textbf{Train Dataset} & 
\multicolumn{1}{c}{\textbf{C-9}} &
\multicolumn{1}{c}{\textbf{C-2 \& C-9}} &
\multicolumn{1}{c}{\textbf{S \& C-9}}  

\\\midrule
Majority-class & 39.1 & - & - \\
SVM (tf-idf) & 78.6 & - & - \\
\citeauthor{preotiuc2019automatically}            &  79.0 & - & - \\
\midrule
XLM \finetuning & 78.6 & 79.3 & 80.1   \\
XLM \pretraining \finetuning & 82.4 & 80.0 & 82.8  \\

\bottomrule
\end{tabular}
\end{center}
\caption{F1 test scores on Complaint-9 for finetuning on the end task alone (indicated as \textbf{C-9}) vs.\ on Complaint-2 or Sanders and again on Complaint-9 (columns \textbf{C-2 \& C-9}, respectively, \textbf{S \& C-9}.).
}\label{tb:task_transferability}
\end{table}

\section{Conclusion}
We investigated multilingual and across-task transfer learning for customer support tasks, based on transformer-based language models. We confirmed prior insights that finetuning the models on low-resource end tasks is important. Additional pretraining on a moderately sized in-domain corpus, however, provides a complementary increase in effectiveness, especially in the non-English setting and starting from a generic multilingual language model. We provide a newly collected multilingual in-domain corpus for customer service tasks and derive the aforementioned findings from experiments using it on five different tasks.

\section{Acknowledgments}
This research received funding from the Flemish Government under the Research Program Artificial Intelligence - 174B09119. We would also like to thank the anonymous reviewers for their valuable and constructive feedback.

\bibliographystyle{acl_natbib}
\bibliography{custom}

\end{document}